\documentclass{article}

\PassOptionsToPackage{numbers, compress}{natbib}

\usepackage[preprint]{neurips_2024} 

\usepackage[utf8]{inputenc} 
\usepackage[T1]{fontenc}    
\usepackage{hyperref}       
\usepackage{url}            
\usepackage{booktabs}       
\usepackage{amsfonts}       
\usepackage{nicefrac}       
\usepackage{microtype}      

\usepackage[dvipsnames]{xcolor}
\definecolor{myRed}{RGB}{213,94,0} 
\definecolor{myGreen}{RGB}{0,158,115} 
\definecolor{myBlue}{RGB}{0,114,178} 
\definecolor{myYellow}{RGB}{240,228,66} 
\definecolor{mySkyBlue}{RGB}{86,180,233} 
\definecolor{myPink}{RGB}{204,121,167} 
\definecolor{myGold}{RGB}{230,159,0} 

\usepackage{color, colortbl}

\usepackage{listings}

\usepackage{pgfplots}
\pgfplotsset{width=7cm, compat=1.9}

\definecolor{LightCyan}{rgb}{0.88,1,1} 
\definecolor{Pink}{rgb}{1, 0.71, 0.756} 
\definecolor{PaleGreen}{rgb}{0.8, 1, 0.8} 
\definecolor{PaleBlue}{rgb}{0.8, 0.9, 1} 
\definecolor{Orange}{RGB}{255, 180, 130} 
\definecolor{Purple}{RGB}{208, 187, 255} 
\definecolor{BrightPink}{RGB}{250, 176, 228} 
\definecolor{Yellow}{RGB}{255, 254, 163} 
\definecolor{Brown}{RGB}{222, 187, 155} 

\usepackage{times}
\usepackage{epsfig}
\usepackage{graphicx}
\usepackage{amsmath}
\usepackage{amssymb}
\usepackage{diagbox}
\usepackage{adjustbox}
\usepackage{xltabular}

\usepackage[symbol]{footmisc}

\usepackage{tikz,lipsum,lmodern}
\usetikzlibrary{shapes.geometric, arrows}
\usepackage[most]{tcolorbox}
\usepackage{caption}
\usepackage{subcaption}

\usepackage{wrapfig}
\usepackage{comment}

\usepackage[textsize=tiny]{todonotes}
\usepackage{blindtext}
\usepackage{multirow}
\usepackage{makecell}

\definecolor{light-gray}{gray}{0.95}

\usepackage{graphicx}
\usepackage{array}
\usepackage{booktabs}
\newcolumntype{N}{>{\centering\arraybackslash}m{.5in}}
\newcolumntype{G}{>{\centering\arraybackslash}m{1in}}
\newcolumntype{H}{>{\centering\arraybackslash}m{1.1in}}
\newcolumntype{M}{>{\centering\arraybackslash}m{.3in}}
\newcolumntype{P}{>{\centering\arraybackslash}m{.8in}}
\newcolumntype{Q}{>{\centering\arraybackslash}m{.4in}}

\begin{document}

\title{Rephrasing natural text data with different languages and quality levels for Large Language Model pre-training}

\author{
Michael Pieler \quad Marco Bellagente \quad Hannah Teufel \quad Duy Phung \\ Nathan Cooper \quad Jonathan Tow \quad Paulo Rocha \quad Reshinth Adithyan \\ Zaid Alyafeai
\quad Nikhil Pinnaparaju \quad Maksym Zhuravinskyi \quad Carlos Riquelme
\\
\\
\textbf{Stability AI Language Team}\thanks{Correspondance to: michael@stability.ai}
}
\maketitle

\renewcommand{\thefootnote}{\arabic{footnote}}

\begin{abstract}
Recently published work on rephrasing natural text data for pre-training LLMs has shown promising results when combining the original dataset with the synthetically rephrased data. We build upon previous work by replicating existing results on C4 and extending them with our optimized rephrasing pipeline to the English, German, Italian, and Spanish Oscar subsets of CulturaX. Our pipeline leads to increased performance on standard evaluation benchmarks in both the mono- and multilingual setup. In addition, we provide a detailed study of our pipeline, investigating the choice of the base dataset and LLM for the rephrasing, as well as the relationship between the model size and the performance after pre-training. By exploring data with different perceived quality levels, we show that gains decrease with higher quality. Furthermore, we find the difference in performance between model families to be bigger than between different model sizes. This highlights the necessity for detailed tests before choosing an LLM to rephrase large amounts of data. Moreover, we investigate the effect of pre-training with synthetic data on supervised fine-tuning. Here, we find increasing but inconclusive results that highly depend on the used benchmark. These results (again) highlight the need for better benchmarking setups. In summary, we show that rephrasing multilingual and low-quality data is a very promising direction to extend LLM pre-training data.
\end{abstract}

\section{Introduction}
The rapid advancements of Large Language Models (LLMs) have transformed them into powerful tools for natural language processing (NLP). While early breakthroughs were often driven by architectural innovations, the field has recently converged toward similar Transformer-based architectures. 
As a result, differences in the training data composition have emerged as a critical differentiator to create high-performance LLMs. Furthermore, as model pre-training continues to be scaled, it has been questioned if natural text data may become scarce \cite{villalobos2024rundatalimitsllm}.

Synthetic data has been proposed to alleviate potential data bottlenecks, while offering increased control over the data properties.
By leveraging a large and diverse set of web-based text, such approaches could mitigate biases from the model or task they are based on. (See related work in section \ref{sec:relwork}.)

In this work, we build upon the text rephrasing from Maini, et al. (2024) \cite{maini2024rephrasing}. Their rephrasing setup leverages LLMs to create large scale pre-training data. Such a rephrasing setup can be interpreted as a form of paraphrasing text augmentation \cite{Li_2022}, that is applied to longer text passages. Those passages are rewritten in different styles by prompting an LLM with specific instructions. As a first step, we focus on replicating the work from Maini, et al. (2024) \cite{maini2024rephrasing} on rephrasing the English Colossal Clean Crawled Corpus (C4) dataset \cite{c4}. We then extend this approach to additional datasets with different languages and varying perceived quality levels. Our contributions can be summarized as follows:
\begin{itemize}
    \item 
    We successfully replicate the rephrasing study by Maini, et al. (2024) \cite{maini2024rephrasing} with a larger LLM trained with double the batch size on C4.
    \item
    We successfully extend our rephrasing pipeline to the English, German, Italian, and Spanish Oscar subsets of CulturaX (CX) \cite{culturax}.
    \item
    In addition, we compare the rephrasing setup on web-based datasets with different perceived quality levels, i.e., CX, C4, and FineWeb-Edu (FWE) \cite{fineweb}.
    \item
    Furthermore, we study the impact of the rephrasing model, by comparing two state-of-the-art LLMs, namely Mistral 7B Instruct v0.2 \cite{jiang2023mistral} and Qwen2 7B Instruct \cite{qwen2} as well as differently sized models from the Qwen2 family.
    \item
    Finally, we apply supervised fine-tuning (SFT) to our LLMs pre-trained on rephrased data to investigate its downstream effect.
\end{itemize}
\section{Related work}\label{sec:relwork}

Curating and selecting data for LLM pre-training is an active area of research, and spans a wide range of approaches and techniques \cite{albalak2024surveydataselectionlanguage, longpre2023pretrainersguidetrainingdata}. 
Different pipelines have shown how to prepare web data into training data \cite{penedo2023refinedwebdatasetfalconllm, wenzek2019ccnetextractinghighquality, penedo2024finewebdatasetsdecantingweb, soldaini2024dolmaopencorpustrillion}, including heuristics based filtering \cite{rae2022scalinglanguagemodelsmethods}, classifiers trained to select content with high educational value \cite{gururangan-etal-2022-whose, penedo2024finewebdatasetsdecantingweb}, and de-duplication \cite{lee-etal-2022-deduplicating, abbas2023semdedupdataefficientlearningwebscale}. 
At the granularity of a fixed, given corpus Sang, et al. (2023) \cite{xie2023doremioptimizingdatamixtures} proposed to compute weights based on a distributionally robust proxy model. Furthermore, a recent setup uses pre-trained open source models to estimate correlations between benchmark scores and documents \cite{thrush2024improvingpretrainingdatausing}.

A large variety of text data augmentation strategies have been described for NLP \cite{Li_2022}. Of particular interest for our work is paraphrasing. Paraphrasing can be carried out at different levels, i.e., starting at a level of words, phrases, sentences, up to text passages, and with different setups, ranging from simpler setups, e.g., Thesauruses- \cite{zhang2016characterlevelconvolutionalnetworkstext} or rule-based \cite{coulombe2018textdataaugmentationsimple, wei-zou-2019-eda}, up to more complex setups, e.g., by using embeddings \cite{mikolov2013distributedrepresentationswordsphrases} or generating text with LLMs \cite{hou2018sequencetosequencedataaugmentationdialogue}.

Recently, generating synthetic training data with LLMs has been increasingly explored. For pre-training data a range of pipelines have been published to generate, e.g., synthetic short stories \cite{eldan2023tinystoriessmalllanguagemodels}, textbooks and exercises \cite{gunasekar2023textbooksneed}, blog posts, stories, posts, and other articles \cite{cosmopedia}, as well as web-based text rephrased in different styles \cite{maini2024rephrasing}. Additionally, several setups have been described for fine-tuning, e.g., back translation of instruction data \cite{li2024selfalignmentinstructionbacktranslation}, creation of alignment data \cite{nvidia2024nemotron4340btechnicalreport, sudalairaj2024lablargescalealignmentchatbots, xu2024magpiealignmentdatasynthesis}, generating text data based on specific taxonomies \cite{li2024syntheticdataalmostscratch}, use cases, skills \cite{xu2023wizardlmempoweringlargelanguage, wang2024codeclmaligninglanguagemodels}, or personas \cite{ge2024scalingsyntheticdatacreation}. Furthermore, synthetic text data can also be useful for inference, e.g., for test-time augmentation \cite{obrien2024improvingblackboxrobustnessincontext}.

In addition, we point the interested reader to more extensive reviews of this topic \cite{liu2024bestpracticeslessonslearned,guo2024generativeaisyntheticdata,long2024llmsdrivensyntheticdatageneration,synthfrontiers}.
\section{Methods}\label{sec:methods}

\subsection{Natural text data}\label{sec:data}
For our experiments, we compare three web-based pre-training datasets: C4 \cite{c4}, the Oscar subsets from CulturaX (CX) \cite{culturax}, and the FineWeb-Edu (FWE) dataset \cite{fineweb}.
These datasets undergo different processing and filtering heuristics, enabling us to validate our method on datasets of varying perceived quality and quantify the potential improvements achieved by rephrasing thoroughly cleaned data All natural text data sources are listed in Table \ref{tab:data-rephrasing} with their document and token counts for the original and the rephrased datasets. The token counts are based on the Stable LM 2 tokenizer \cite{bellagente2024stable}. The Ask-LLM \cite{askllm} subsets are filtered with a score threshold of >0.6 (A>0.6) and >0.97 (A>0.97), and the FWE classifier subsets are filtered with a score threshold of >2.5 (F>2.5; for more details about the classifiers see Appendix \ref{sec:dataclassifier}).
\begin{table}[ht]
\caption{Overview of all original and rephrased (re.) datasets. C4, CulturaX (CX), and FineWeb-Edu (FWE) are used as base datasets. The default rephrasing model is Mistral 7B Instruct v0.2, except for the Qwen2 Instruct model size experiments. The classifier filtered datasets are indicated with the corresponding threshold, e.g., Ask-LLM > 0.6 (A>0.6), FWE > 2.5 (F>2.5). The token counts are based on the Stable LM 2 tokenizer.}
\centering
\noindent
\begin{tabular}{l r r r r r}
\toprule
\multicolumn{1}{ l }{\textbf{Dataset}} & \textbf{mio. docs} & \textbf{B tokens} \\[0.5ex] 
\midrule
\midrule

\multicolumn{1}{ l }{\textbf{C4 (English)}} & 365 & 172 \\ [0.2ex]
\multicolumn{1}{ l }{\textbf{\: \rotatebox[origin=c]{180}{$\Lsh$} C4 A>0.6}} & 34 & 20 \\ [0.2ex]
\midrule
\multicolumn{1}{ l }{\textbf{C4 rephrased toddler }} & 343 & 85 \\ [0.2ex]
\multicolumn{1}{ l }{\textbf{\: \rotatebox[origin=c]{180}{$\Lsh$} C4 rephrased toddler  A>0.6}} & 336 & 84 \\ [0.2ex]
\multicolumn{1}{ l }{\textbf{\quad \rotatebox[origin=c]{180}{$\Lsh$} C4 rephrased toddler A>0.97}} & 33 & 5 \\ [0.2ex]
\midrule

\multicolumn{1}{ l }{\textbf{C4 rephrased hard }} & 341 & 167 \\ [0.2ex]
\multicolumn{1}{ l }{\textbf{\: \rotatebox[origin=c]{180}{$\Lsh$} C4 rephrased hard A>0.6}} & 338 & 160 \\ [0.2ex]
\multicolumn{1}{ l }{\textbf{\quad \rotatebox[origin=c]{180}{$\Lsh$} C4 rephrased hard A>0.97}} & 38 & 18 \\ [0.2ex]
\midrule

\multicolumn{1}{ l }{\textbf{C4 rephrased wiki }} & 334 & 161 \\ [0.2ex]
\multicolumn{1}{ l }{\textbf{\: \rotatebox[origin=c]{180}{$\Lsh$} C4 rephrased wiki A>0.6}} & 332 & 152 \\ [0.2ex]
\multicolumn{1}{ l }{\textbf{\quad \rotatebox[origin=c]{180}{$\Lsh$} C4 rephrased wiki A>0.97}} & 52 & 16 \\ [0.2ex]
\midrule

\multicolumn{1}{ l }{\textbf{C4 rephrased QA }} & 299 & 176 \\ [0.2ex]
\multicolumn{1}{ l }{\textbf{\: \rotatebox[origin=c]{180}{$\Lsh$} C4 rephrased QA A>0.6}} & 298 & 162 \\ [0.2ex]
\multicolumn{1}{ l }{\textbf{\quad \rotatebox[origin=c]{180}{$\Lsh$} C4 rephrased QA A>0.97}} & 23 & 9 \\ [0.2ex]
%
\midrule
\midrule

\multicolumn{1}{ l }{\textbf{CX-E (English)}} & 843 & 741 \\ [0.2ex]
\multicolumn{1}{ l }{\textbf{CX-E rephrased toddler}} & 227 & 93 \\ [0.2ex]
\multicolumn{1}{ l }{\textbf{CX-E rephrased wiki}} & 170 & 130 \\ [0.2ex]
\multicolumn{1}{ l }{\textbf{CX-E rephrased QA}} & 157 & 137 \\ [0.2ex]
\midrule
\multicolumn{1}{ l }{\textbf{CX-E Qwen2 0.5B Inst. re. QA opt.}} & 104 & 73 \\ [0.2ex]
\multicolumn{1}{ l }{\textbf{CX-E Qwen2 1.5B Inst. re. QA opt.}} & 107 & 84 \\ [0.2ex]
\multicolumn{1}{ l }{\textbf{CX-E Qwen2 7B Inst. re. QA opt.}} & 105 & 102 \\ [0.2ex]
\midrule

\multicolumn{1}{ l }{\textbf{CX-G (German)}} & 105 & 96 \\ [0.2ex]
\multicolumn{1}{ l }{\textbf{CX-G rephrased QA optimized}} & 85 & 56 \\ [0.2ex]
\midrule

\multicolumn{1}{ l }{\textbf{CX-S (Spanish)}} & 86 & 78 \\ [0.2ex]
\multicolumn{1}{ l }{\textbf{CX-S rephrased QA optimized}} & 75 & 51 \\ [0.2ex]
\midrule

\multicolumn{1}{ l }{\textbf{CX-I (Italian)}} & 46 & 48 \\ [0.2ex]
\multicolumn{1}{ l }{\textbf{CX-I rephrased QA optimized}} & 38 & 28 \\ [0.2ex]
\midrule
\midrule

\multicolumn{1}{ l }{\textbf{FWE}} & 97 & 100 \\ [0.2ex]
\multicolumn{1}{ l }{\textbf{FWE rephrased QA optimized}} & 95 & 98 \\ [0.2ex]
\multicolumn{1}{ l }{\textbf{\quad \rotatebox[origin=c]{180}{$\Lsh$} FWE re. QA opt. F>2.5}} & 74 & 79 \\ [0.2ex]
\midrule
\midrule

\multicolumn{1}{ l }{\textbf{Total original}} & 1,542 & 1,234 \\ [0.2ex]
\multicolumn{1}{ l }{\textbf{Total unique rephrased}} & 2,777 & 1,630 \\ [0.2ex]
\bottomrule
\end{tabular}
\label{tab:data-rephrasing}
\end{table}

\subsection{Rephrasing natural text data}\label{sec:rephrase_nat}
The rephrasing pipeline consists of several steps: In the preprocessing step, we first split the raw text into smaller passages (section \ref{sec:preprocessing}). These passages are then inserted into prompt templates (section \ref{sec:prompt_templates}) for model inference to obtain the rephrased text (section \ref{sec:inference}). The rephrased passages are cleaned and merged back into full documents again in the postprocessing step (section \ref{sec:postprocessing}). Finally, the rephrased data is used for the LLM pre-training (Appendix \ref{sec:traincfg}; pipeline overview in Appendix \ref{sec:pipeline_overview}).

\subsubsection{Preprocessing}\label{sec:preprocessing}
The preprocessing splits the documents into smaller passages. We extend the passage length of 300 tokens from Maini, et al. (2024) \cite{maini2024rephrasing} to 350 tokens and introduce a minimum passage length of 50 tokens. We increased the token limit to 350 due to our merge logic and did not observe any degradation in our rephrasing experiments.
For passage generation, we use a simple and fast preprocessing algorithm:
\begin{enumerate}
  \item Generate smaller passages, by splitting documents on line breaks, i.e., \texttt{\textbackslash n}.
  \item Remove empty passages.
  \item Split passages exceeding 350 tokens on common sentence-ending characters, i.e., \texttt{.}, \texttt{!}, or \texttt{?} followed by one or more white space characters.
  \item Merge consecutive passages until we reach the maximum target length of 350 tokens. If the total passage length is longer than 350 tokens after a merge we start a new merge process with the last passage.
\end{enumerate}
To obtain the token counts in the preprocessing step, we use a token count per character estimate based on a random subset of the data to avoid running tokenization every time we want to verify the passage length. Compared to the pre-processing setup described by Maini, et al. (2024) \cite{maini2024rephrasing}, we use simple regex patterns instead of the NLTK sentence splitter and omit the check for specific delimiters in the text. However, our results on C4 are better than theirs and are discussed in detail in section \ref{sec:c4}.

\subsubsection{Prompt templates}\label{sec:prompt_templates}
In this work we use two different sets of prompt templates. First, for replicating the previous work on C4 (section \ref{sec:c4}) and comparing to the CX English (CX-E) dataset (section \ref{sec:cultx}), we use the toddler, hard, wiki, and QA prompt templates from Maini, et al. (2024) \cite{maini2024rephrasing}. We adapted the toddler, hard, wiki, and QA prompt templates after initial experiments with the newer model version v0.2 of the Mistral 7B Instruct model.
Then, we create new prompt templates, optimized for easier postprocessing, which are applied in the experiments on the FWE dataset in the data quality investigations (section \ref{sec:dataquality}) and the model scale ablations (section \ref{sec:modelscales}). 
Finally, we translate those prompt templates to other languages, i.e., German, Spanish, and Italian, for our multilingual experiments (section \ref{sec:cultx}). All prompt templates are shown in Appendix \ref{sec:prompts_rephr}.

\subsubsection{Inference}\label{sec:inference}
We rely on Mistral 7B Instruct v0.2 \cite{jiang2023mistral} for inference, in contrast to the setup used by Maini, et al.(2024) \cite{maini2024rephrasing}, which used Mistral 7B Instruct v0.1. In order to investigate the different performance of similarly sized models, as well as differently sized models within the same family, we additionally use the Qwen2 0.5B, 1.5B, and 7B Instruct models \cite{qwen2} for QA rephrasing the CX-E dataset in section \ref{sec:modelscales}.

\subsubsection{Postprocessing}\label{sec:postprocessing}
In the postprocessing, we apply two different transformations based on whether the outputs were generated with prompts following Maini, et al. (2024) \cite{maini2024rephrasing} or our optimized prompt templates for simple text extraction. Below are the postprocessing steps for the outputs generated using prompts based on Maini, et al. (2024) \cite{maini2024rephrasing}:
\begin{enumerate}
  \item Identify and split multiple paraphrases within the output and randomly return one.
  \item Remove any unwanted elements in the output, such as \texttt{Paraphrase:}, \texttt{Toddler-friendly paraphrase:}, \texttt{Erudite paraphrase:}, and other  patterns.
  \item Only keep passages between 50 and 5,000 characters.
  \item Discard passages if the last character is alphabetic, as this indicates a truncated output.
  \item Merge passages to the full documents again.
  \item Remove documents with fewer than 100 characters.
\end{enumerate}

For outputs generated with our optimized prompt setup, we replace step 1 and 2 from above by extracting the content between the first \texttt{<text></text>} tag pair, which is a much simpler and more efficient setup.
\section{Results}\label{sec:results}
We evaluate our pre-training experiments with the Language Model Evaluation Harness \cite{eval-harness} (MIT License) on the following established natural language benchmarks: ARC Challenge (A-C), ARC Easy (A-E) \cite{arc}, HellaSwag (HS) \cite{hellaswag}, Lambada (L) \cite{lambada}, PIQA (P) \cite{piqa}, SciQ (SQ) \cite{sciq}, and WinoGrande (WG) \cite{winogrande}. For the evaluation of our fine-tuned models, we use the Open LLM 1 and 2 benchmarks \cite{open-llm-leaderboard-v1,open-llm-leaderboard-v2}

\subsection{Rephrasing C4}\label{sec:c4}
First, we rephrase C4 to verify that we see similar results with our experimental setup compared to previous work by Maini, et al. (2024) \cite{maini2024rephrasing}. Table \ref{tab:rephrasing-comparison} shows our C4 rephrasing experiments grouped together with the most similar experimental setup from Maini, et al. (2024) \cite{maini2024rephrasing}. Our setup outperforms the previous work between 0.2 and 2.5 percentage points. Interestingly, we observe larger, systematic differences at the benchmark level. In our experiments, scores on A-C and HS are always higher, whereas on A-E they are always lower.
Those differences can be very likely linked to the different training setups and models used for rephrasing the data (see Appendix \ref{sec:traincfg} and section \ref{sec:inference}).

\begin{table}[ht]
\caption{Replicated C4 experiments with the baseline and rephrased (re.) data. Experiments are grouped together with the most similar experiment from Maini, et al. (2024) \cite{maini2024rephrasing}. Each section uses the experiment from Maini, et al. (2024) 
 \cite{maini2024rephrasing} as baseline to indicate the change. The used rephrasing prompt is indicated in the dataset name with hard, QA, toddler (tod.), and wiki.}
\centering
\noindent
\begin{tabular}{l r r r r r r l}
\toprule

\multicolumn{1}{ l }{\textbf{Dataset}} & A-C & A-E & HS & P & SQ & WG & Avg \\ [0.5ex]
\midrule
\midrule


\multicolumn{1}{ l }{\textbf{"Full C4 (170B)"}} & 26.8 & 61.6 & 46.8 & 74.9 & 85.0 & 59.0 & 59.0 \\ [0.2ex]

\multicolumn{1}{ l }{\textbf{C4}} & 28.0 & 51.7 & 58.7 & 73.3 & 86.2 & 57.4 & 59.2 {\color{myGreen}{+0.2}}\\ [0.2ex]


\midrule






\multicolumn{1}{ l }{\textbf{"Synthetic+C4 (85B)"}} & 29.9 & 64.1 & 46.2 & 75.4 & 87.6 & 58.9 & 60.4 \\ [0.2ex]

\multicolumn{1}{ l }{\textbf{C4 1:1 re. hard QA tod. wiki}} & 31.1 & 52.4 & 59.8 & 73.9 & 87.5 & 59.1 & 60.6 {\color{myGreen}{+0.2}}\\ [0.2ex]


\midrule

\multicolumn{1}{ l }{\textbf{"Med-35B"} (wiki)} & 27.0 & 56.6 & 41.9 & 74.0 & 80.0 & 53.4 & 55.5 \\ [0.2ex]

\multicolumn{1}{ l }{\textbf{C4 re. wiki}} & 31.7 & 48.7 & 51.7 & 72.4 & 82.9 & 57.5 & 57.5 {\color{myGreen}{+2.0}}\\ [0.2ex]

\midrule

\multicolumn{1}{ l }{\textbf{"QA-35B"}} & 27.1 & 61.7 & 43.4 & 75.2 & 85.5 & 53.9 & 57.8 \\ [0.2ex]

\multicolumn{1}{ l }{\textbf{C4 re. QA}} & 31.7 & 52.1 & 54.7 & 72.9 & 83.8 & 57.3 & 58.7 {\color{myGreen}{+0.9}}\\ [0.2ex]

\midrule

\multicolumn{1}{ l }{\textbf{"QA+C4-35B"}} & 29.0 & 62.2 & 44.6 & 74.8 & 85.1 & 55.7 & 58.6 \\ [0.2ex]

\multicolumn{1}{ l }{\textbf{C4 1:1 re. QA}} & 32.6 & 53.1 & 60.1 & 74.2 & 89.2 & 57.1 & 61.1 {\color{myGreen}{+2.5}}\\ [0.2ex]



\midrule

\multicolumn{1}{ l }{\textbf{"Combined-1:1-35B"} (QA, wiki)} & 28.2 & 60.6 & 43.7 & 73.8 & 85.9 & 57.7 & 58.3 \\ [0.2ex]


\multicolumn{1}{ l }{\textbf{C4 1:1 re. QA tod. wiki}} & 30.9 & 51.8 & 59.9 & 73.8 & 88.6 & 57.9 & 60.5 {\color{myGreen}{+2.2}}\\ [0.2ex]


\bottomrule
\end{tabular}
\label{tab:rephrasing-comparison}
\end{table}

Our full range of C4 experiments is shown in Table \ref{tab:rephrasing-c4}. Models trained on only the rephrased data consistently show a benchmarking average below our C4 baseline, which ranges from -5.5 percentage points for the hard rephrased data to -1.8 percentage points for the toddler rephrased data. The Ask-LLM filtering did not improve our baseline and makes most of the rephrased datasets worse. In contrast, when the original C4 dataset is 1:1 mixed with the hard, QA, toddler, and wiki rephrased data, we see improvements of up to +1.7 percentage points. Previous work has shown similar results for the interaction between synthetic and original data and highlighted the need for mixing both data types \cite{alemohammad2023selfconsuminggenerativemodelsmad,gerstgrasser2024modelcollapseinevitablebreaking}. However, we don't observe the same trend for all data we investigated, as shown in the following sections. No improvements could be achieved for the original C4 data mixed with the Ask-LLM filtered QA rephrased data with a score threshold of >0.6 and >0.97. Because of the strong QA rephrasing performance, which was also shown by Maini, et al. (2024) \cite{maini2024rephrasing}, we continue to use the QA setup for our other experiments.
\begin{table}[ht]
\caption{All C4 experiments with the baseline and rephrased (re.) data. The change to the C4 baseline is indicated. Ask-LLM classifier filtering is shown with the used threshold, e.g., A>0.6 or A>0.97. In the last section the base and the rephrased data is 1:1 mixed.}
\centering
\noindent
\begin{tabular}{l r r r r r r r l}
\toprule

\multicolumn{1}{ l }{\textbf{Dataset}} & A-C & A-E & HS & L & P & SQ  & WG & Avg \\ [0.5ex]
\midrule
\midrule

\multicolumn{1}{ l }{\textbf{C4 (baseline)}} & 28.0 & 51.7 & 58.7 & 48.2 & 73.3 & 86.2 & 57.4 & 57.6 \\ [0.2ex]

\multicolumn{1}{ l }{\textbf{C4 A>0.6}} & 29.2 & 52.8 & 58.0 & 48.1 & 72.7 & 85.7 & 55.6 & 57.4 {\color{myRed}{-0.2}}\\ [0.2ex]

\midrule

\multicolumn{1}{ l }{\textbf{C4 re. toddler}} & 28.4 & 49.1 & 50.0 & 42.6 & 72.2 & 87.9 & 60.6 & 55.8 {\color{myRed}{-1.8}}\\ [0.2ex]

\multicolumn{1}{ l }{\textbf{C4 re. toddler A>0.6}} & 27.4 & 49.1 & 50.0 & 40.7 & 72.1 & 87.5 & 58.6 & 55.1 {\color{myRed}{-2.5}}\\ [0.2ex]

\multicolumn{1}{ l }{\textbf{C4 re. toddler A>0.97}} & 26.7 & 41.8 & 42.4 & 27.2 & 67.8 & 77.3 & 54.8 & 48.3 {\color{myRed}{-9.3}}\\ [0.2ex]

\midrule

\multicolumn{1}{ l }{\textbf{C4 re. hard}} & 28.5 & 45.3 & 46.4 & 39.5 & 70.0 & 81.5 & 53.5 & 52.1 {\color{myRed}{-5.5}}\\ [0.2ex]

\multicolumn{1}{ l }{\textbf{C4 re. hard A>0.6}} & 28.4 & 46.1 & 46.2 & 37.3 & 68.9 & 83.1 & 53.6 & 51.9 {\color{myRed}{-5.7}}\\ [0.2ex]

\multicolumn{1}{ l }{\textbf{C4 re. hard A>0.97}} & 25.5 & 41.8 & 41.8 & 28.9 & 66.4 & 78.1 & 49.6 & 47.4 {\color{myRed}{-10.2}}\\ [0.2ex]

\midrule

\multicolumn{1}{ l }{\textbf{C4 re. QA}} & 31.7 & 52.1 & 54.7 & 37.3 & 72.9 & 83.8 & 57.3 & 55.7 {\color{myRed}{-1.9}}\\ [0.2ex]

\multicolumn{1}{ l }{\textbf{C4 re. QA A>0.6}} & 31.2 & 52.9 & 55.4 & 40.9 & 73.6 & 84.8 & 57.1 & 56.6 {\color{myRed}{-1.0}}\\ [0.2ex]

\multicolumn{1}{ l }{\textbf{C4 re. QA A>0.97}} & 28.5 & 45.7 & 47.8 & 31.2 & 71.3 & 81.5 & 54.1 & 51.4 {\color{myRed}{-6.2}}\\ [0.2ex]

\midrule

\multicolumn{1}{ l }{\textbf{C4 re. wiki}} & 31.7 & 48.7 & 51.7 & 43.3 & 72.4 & 82.9 & 57.5 & 55.4 {\color{myRed}{-2.2}}\\ [0.2ex]


\midrule

\multicolumn{1}{ l }{\textbf{C4 1:1 re. all}} & 31.1 & 52.4 & 59.8 & 51.1 & 73.9 & 87.5 & 59.1 & 59.3 {\color{myGreen}{+1.7}}\\ [0.2ex]

\multicolumn{1}{ l }{\textbf{C4 1:1 re. QA tod. wiki}} & 30.9 & 51.8 & 59.9 & 50.3 & 73.8 & 88.6 & 57.9 & 59.0 {\color{myGreen}{+1.4}}\\ [0.2ex]

\multicolumn{1}{ l }{\textbf{C4 1:1 re. QA}} & 32.6 & 53.1 & 60.1 & 49.0 & 74.2 & 89.2 & 57.1 & 59.3 {\color{myGreen}{+1.7}}\\ [0.2ex]

\multicolumn{1}{ l }{\textbf{C4 1:1 re. QA A>0.6}} & 30.7 & 53.1 & 60.3 & 48.4 & 73.3 & 88.7 & 59.0 & 59.1 {\color{myGreen}{+1.5}}\\ [0.2ex]

\multicolumn{1}{ l }{\textbf{C4 1:1 re. QA A>0.97}} & 30.0 & 51.8 & 58.3 & 48.8 & 74.6 & 87.6 & 57.9 & 58.4 {\color{myGreen}{+0.8}}\\ [0.2ex]
\bottomrule
\end{tabular}
\label{tab:rephrasing-c4}
\end{table}

\subsection{Rephrasing multilingual CulturaX}\label{sec:cultx}
To verify if the rephrasing works on multilingual data, we apply the QA rephrasing on the English, German, Spanish, and Italian Oscar subsets of CulturaX (CX-E/G/S/I) \cite{culturax}. Table \ref{tab:rephrasing-cx} shows the multilingual CX rephrasing results. Only the CX-E rephrased QA data shows a slightly worse result with an absolute difference of -0.2 percentage points compared to the baseline. All other experiments with the QA rephrased data and 1:1 mixed data show higher gains between +1.5 to +3.7 percentage points. In particular, the German, Spanish, and Italian rephrased data and the corresponding 1:1 mixed data even show improvements larger than +3.1 percentage points. Those large increases are in contrast to the C4 rephrasing results in Table \ref{tab:rephrasing-c4} and could be explained by the lower perceived quality of CX, especially in languages other than English. For such low-quality datasets, the rephrasing seems to improve the base dataset by a big margin.
\begin{table}[ht]
\caption{Multilingual CulturaX (CX) experiments with the baseline and rephrased (re.) data. Experiments are grouped together based on the language, i.e, English (E), German (G), Spanish (G), Italian (I), or all combined (all). The change to the corresponding baseline is indicated. In the mixed experiments the base and the rephrased data are combined 1:1.}
\centering
\noindent
\begin{tabular}{l r r r r r r r l}
\toprule

\multicolumn{1}{ l }{\textbf{Dataset}} & A-C & A-E & HS & L & P & SQ & WG & Avg \\ [0.5ex]
\midrule
\midrule


\multicolumn{1}{ l }{\textbf{CX-E (baseline)}} & 28.9 & 50.2 & 54.9 & 45.7 & 73.2 & 85.8 & 56.0 & 56.4 \\ [0.2ex]

\multicolumn{1}{ l }{\textbf{CX-E re. QA}} & 32.2 & 50.5 & 54.0 & 39.8 & 72.6 & 84.4 & 59.6 & 56.2 {\color{myRed}{-0.2}}\\ [0.2ex]

\multicolumn{1}{ l }{\textbf{CX-E 1:1 re. QA}} & 32.3 & 50.1 & 56.7 & 49.0 & 73.7 & 86.3 & 57.9 & 58.0 {\color{myGreen}{+1.6}}\\ [0.2ex]




\midrule
\multicolumn{1}{ l }{\textbf{CX-G (baseline)}} & 26.3 & - & 39.5 & 44.7 & - & - & - & 36.8 \\ [0.2ex]

\multicolumn{1}{ l }{\textbf{CX-G re. QA}} & 29.3 & - & 41.1 & 49.8 & - & - & - & 40.0 {\color{myGreen}{+3.2}}\\ [0.2ex]

\multicolumn{1}{ l }{\textbf{CX-G 1:1 re. QA}} & 29.4 & - & 40.4 & 51.7 & - & - & - & 40.5 {\color{myGreen}{+3.7}}\\ [0.2ex]

\midrule
\multicolumn{1}{ l }{\textbf{CX-S (baseline)}} & 27.7 & - & 44.7 & 28.5 & - & - & - & 33.6 \\ [0.2ex]
\multicolumn{1}{ l }{\textbf{CX-S re. QA}} & 28.8 & - & 46.7 & 35.6 & - & - & - & 37.0 {\color{myGreen}{+3.4}}\\ [0.2ex]
\multicolumn{1}{ l }{\textbf{CX-S 1:1 re. QA}} & 28.2 & - & 47.0 & 34.8 & - & - & - & 36.7 {\color{myGreen}{+3.1}}\\ [0.2ex]

\midrule
\multicolumn{1}{ l }{\textbf{CX-I (baseline)}} & 26.0 & - & 40.2 & 30.3 & - & - & - & 32.2 \\ [0.2ex]
\multicolumn{1}{ l }{\textbf{CX-I re. QA}} & 28.6 & - & 42.4 & 35.6 & - & - & - & 35.5 {\color{myGreen}{+3.3}}\\ [0.2ex]
\multicolumn{1}{ l }{\textbf{CX-I 1:1 re. QA}} & 28.7 & - & 42.4 & 35.6 & - & - & - & 35.6 {\color{myGreen}{+3.4}}\\ [0.2ex]

\midrule

\multicolumn{1}{ l }{\textbf{CX-all (baseline E})} & 27.4 & 47.3 & 45.7 & 37.4 & 69.4 & 80.8 & 52.3 & 51.5 \\ [0.2ex]
\multicolumn{1}{ l }{\textbf{CX-all (baseline G)}} & 26.2 & - & 39.0 & 44.6 & - & - & - & 36.6 \\ [0.2ex]
\multicolumn{1}{ l }{\textbf{CX-all (baseline S)}} & 25.9 & - & 43.4 & 29.2 & - & - & - & 32.8 \\ [0.2ex]
\multicolumn{1}{ l }{\textbf{CX-all (baseline I)}} & 25.7 & - & 41.4 & 32.6 & - & - & - & 33.2 \\ [0.2ex]

\multicolumn{1}{ l }{\textbf{CX-all 1:1 re. QA (E)}} & 28.2 & 46.0 & 47.3 & 41.7 & 69.6 & 83.2 & 54.7 & 53.0 {\color{myGreen}{+1.5}}\\ [0.2ex]
\multicolumn{1}{ l }{\textbf{CX-all 1:1 re. QA (G)}} & 29.0 & - & 40.4 & 46.6 & - & - & - & 38.7 {\color{myGreen}{+2.1}}\\ [0.2ex]
\multicolumn{1}{ l }{\textbf{CX-all 1:1 re. QA (S)}} & 26.3 & - & 44.7 & 32.8 & - & - & - & 34.6 {\color{myGreen}{+1.8}}\\ [0.2ex]
\multicolumn{1}{ l }{\textbf{CX-all 1:1 re. QA (I)}} & 28.1 & - & 42.7 & 36.3 & - & - & - & 35.7 {\color{myGreen}{+2.5}}\\ [0.2ex]


\bottomrule
\end{tabular}
\label{tab:rephrasing-cx}
\end{table}

\subsection{Rephrasing datasets with different quality levels}\label{sec:dataquality}
Based on the CX results in section \ref{sec:cultx} and the potential link to the perceived data quality, we QA rephrase the recently published FineWeb-Edu (FWE) dataset \cite{fineweb} with our optimized prompt setup. FWE is a dataset with an optimized filtering pipeline that includes a classifier filtering step to increase the dataset quality. The perceived quality ranking of the datasets after manual inspection is CX<C4<FWE. The results are shown in Table \ref{tab:rephrasing-dataquality}. Interestingly, the FWE QA rephrased data shows a similar decrease as the C4 QA rephrased data with -1.7 percentage points. In contrast, the FWE data 1:1 mixed with the QA rephrased data shows no improvements but a reduced performance with -0.1 percentage points. Even an additional FWE classifier filtering step with a threshold of >2.5 couldn't improve the results. This is similar to the results with C4 and the Ask-LLM classifier filtering from section \ref{sec:c4}, where also no improvements could be obtained. The CX and C4 1:1 rephrased QA experiments show similar improvements over the baseline with +1.6 and +1.7 percentage points, respectively. Thus, it seems that rephrasing with our setup only benefits low- to medium-quality data when mixed with it.
\begin{table}[ht]
\caption{C4, CX-E, and FWE experiments with the baseline and rephrased (re.) data. Changes to the corresponding baseline are indicated in each section. The classifier filtered datasets are shown with the corresponding threshold, e.g., Ask-LLM > 0.6 (A>0.6), FWE > 2.5 (F>2.5). In the mixed experiments the base and the rephrased data are combined 1:1.}
\centering
\noindent
\begin{tabular}{l r r r r r r r l l}
\toprule

\multicolumn{1}{ l }{\textbf{Dataset}} & A-C & A-E & HS & L & P & SQ & WG & Avg\\ [0.5ex]
\midrule
\midrule





\multicolumn{1}{ l }{\textbf{CX-E (baseline)}} & 28.9 & 50.2 & 54.9 & 45.7 & 73.2 & 85.8 & 56.0 & 56.4 \\ [0.2ex]

\multicolumn{1}{ l }{\textbf{CX-E re. QA}} & 32.2 & 50.5 & 54.0 & 39.8 & 72.6 & 84.4 & 59.6 & 56.2 {\color{myRed}{-0.2}}\\ [0.2ex]

\multicolumn{1}{ l }{\textbf{CX-E 1:1 re. QA}} & 32.3 & 50.1 & 56.7 & 49.0 & 73.7 & 86.3 & 57.9 & 58.0 {\color{myGreen}{+1.6}}\\ [0.2ex]

\midrule



\multicolumn{1}{ l }{\textbf{C4 (baseline)}} & 28.0 & 51.7 & 58.7 & 48.2 & 73.3 & 86.2 & 57.4 & 57.6 \\ [0.2ex]

\multicolumn{1}{ l }{\textbf{C4 re. QA}} & 31.7 & 52.1 & 54.7 & 37.3 & 72.9 & 83.8 & 57.3 & 55.7 {\color{myRed}{-1.9}}\\ [0.2ex]

\multicolumn{1}{ l }{\textbf{C4 1:1 re. QA}} & 32.6 & 53.1 & 60.1 & 49.0 & 74.2 & 89.2 & 57.1 & 59.3 {\color{myGreen}{+1.7}}\\ [0.2ex]

\multicolumn{1}{ l }{\textbf{C4 1:1 re. QA A>0.6}} & 30.7 & 53.1 & 60.3 & 48.4 & 73.3 & 88.7 & 59.0 & 59.1 {\color{myGreen}{+1.5}}\\ [0.2ex]

\multicolumn{1}{ l }{\textbf{C4 1:1 re. QA A>0.97}} & 30.0 & 51.8 & 58.3 & 48.8 & 74.6 & 87.6 & 57.9 & 58.4 {\color{myGreen}{+0.8}}\\ [0.2ex]

\midrule




\multicolumn{1}{ l }{\textbf{FWE (baseline)}} & 37.5 & 64.4 & 55.8 & 44.4 & 71.8 & 90.1 & 58.8 & 60.4 \\ [0.2ex]

\multicolumn{1}{ l }{\textbf{FWE re. QA}} & 38.6 & 61.8 & 51.8 & 42.3 & 70.7 & 89.3 & 56.4 & 58.7 {\color{myRed}{-1.7}}\\ [0.2ex]

\multicolumn{1}{ l }{\textbf{FWE 1:1 re. QA}} & 38.9 & 60.6 & 55.9 & 46.5 & 71.1 & 91.4 & 57.9 & 60.3 {\color{myRed}{-0.1}}\\ [0.2ex]

\multicolumn{1}{ l }{\textbf{FWE 1:1 re. QA F>2.5}} & 39.2 & 61.5 & 55.7 & 45.8 & 72.0 & 91.4 & 57.1 & 60.4 {\color{myGreen}{+0.0}}\\ [0.2ex]

\bottomrule
\end{tabular}
\label{tab:rephrasing-dataquality}
\end{table}

\subsection{Different rephrasing model scales}\label{sec:modelscales}
To better understand the scaling behaviour of the rephrasing model, we used the Qwen2 (Q2) model series \cite{qwen2} that comes with a range of differently-sized LLMs. In particular, we used their 0.5B, 1.5B, and 7B Instruct models for QA rephrasing CX-E. The results are shown in Table \ref{tab:rephrasing-diff-models}. Interestingly, there is no clear trend for the investigated Q2 model scales as the best performing LLM is the one with 1.5B parameters, followed by the smallest one with 0.5B, and the 7B with the worst performance. In addition, our standard rephrasing setup with Mistral 7B Instruct v0.2 shows a higher performance of +1.0 percentage points when compared to Q2 7B Instruct and +0.6 percentage points when compared to Q2 1.5B Instruct. These inconclusive results indicate that model scale alone is insufficient to pick an effective rephrasing model for natural text.
\begin{table}[ht]\caption{Experiments on CX-E QA rephrased (re.) data from different rephrasing models mixed 1:1 with the base data. The middle section uses Mistral 7B Instruct v0.2 and the Qwen2 Instruct model family (Q2) is used in the bottom section. Indicated are the changes to the baseline in the top row.}
\centering
\noindent
\begin{tabular}{l r r r r r r r l}
\toprule

\multicolumn{1}{ l }{\textbf{Dataset}} & A-C & A-E & HS & L & P & SQ & WG & Avg \\ [0.5ex]
\midrule
\midrule

\multicolumn{1}{ l }{\textbf{CX-E (baseline)}} & 28.9 & 50.2 & 54.9 & 45.7 & 73.2 & 85.8 & 56.0 & 56.4 \\ [0.2ex]
\midrule
\multicolumn{1}{ l }{\textbf{CX-E 1:1 re. QA}} & 32.3 & 50.1 & 56.7 & 49.0 & 73.7 & 86.3 & 57.9 & 58.0 {\color{myGreen}{+1.6}}\\ [0.2ex]

\midrule
\multicolumn{1}{ l }{\textbf{CX-E 1:1 Q2 0.5B re. QA}} & 28.8 & 52.7 & 56.2 & 46.4 & 72.6 & 86.3 & 56.5 & 57.1 {\color{myGreen}{+0.7}}\\ [0.2ex]

\multicolumn{1}{ l }{\textbf{CX-E 1:1 Q2 1.5B re. QA}} & 30.1 & 54.1 & 55.8 & 45.9 & 73.3 & 85.6 & 57.2 & 57.4 {\color{myGreen}{+1.0}}\\ [0.2ex]

\multicolumn{1}{ l }{\textbf{CX-E 1:1 Q2 7B re. QA}} & 29.4 & 50.8 & 56.1 & 47.6 & 72.2 & 86.0 & 57.1 & 57.0 {\color{myGreen}{+0.6}}\\ [0.2ex]

\bottomrule
\end{tabular}
\label{tab:rephrasing-diff-models}
\end{table}

\subsection{Supervised fine-tuning}\label{sec:sftresults}
Supervised fine-tuning (SFT) of our pre-trained LLMs on the CX-E, C4, and FWE dataset with and without the corresponding QA rephrased subsets were carried out on UltraChat 200k \cite{ding2023enhancing} to investigate the effect of the QA rephrased data in the pre-training stage. The obtained results for the Open LLM 1 and Open LLM 2 benchmarks are shown in Table \ref{tab:openllm1} and \ref{tab:openllm2}, respectively. On the Open LLM 1 benchmark, all three datasets show an increase with the QA rephrased data in the range of +0.5 to +0.8 percentage points. The FWE baseline and the FWE 1:1 rephrased QA fine-tuning experiment show the highest benchmark average with 39.8\% and 40.5\%, respectively. The benchmark result order is for the baseline and the 1:1 rephrased QA experiments the same, i.e., CX-E<C4<FWE. In contrast, the Open LLM 2 benchmarks show a different picture, with the CX-E experiments showing the highest benchmark averages and no difference between the baseline and the 1:1 rephrased QA experiment. The C4 and the FWE 1:1 rephrased QA experiments are 1.7 and 0.1 percentage points below the baseline, respectively. This results in a different benchmark order when compared to the Open LLM 1 results, with FWE<CX-E<C4 for the baselines and C4<FWE<CX-E for the 1:1 rephrased QA experiments. Interestingly, our results indicate that we don't train on the "test task" \cite{dominguezolmedo2024trainingtesttaskconfounds} by using our QA rephrased data as the results still show gaps between the baseline and the 1:1 mixed QA rephrased experiments after the fine-tuning in most of the cases. Overall, those experiments show that depending on the benchmarking, mixing in QA rephrased data can be beneficial. However, the increases depend highly on the benchmark suite and baseline dataset used.
\begin{table}[ht]
\caption{Open LLM 1 benchmarks of the fine-tuning experiments with the LLMs pre-trained on the baseline data and the baseline data mixed 1:1 with the QA rephrased (re.) version.}
\centering
\noindent
\begin{tabular}{l r r r r l}
\toprule

\multicolumn{1}{ l }{\textbf{Dataset}} & ARC-C & GSM8K & HS & WG & Avg \\ [0.5ex]
\midrule
\midrule

\multicolumn{1}{ l }{\textbf{CX-E (baseline)}} & 32.9 & 1.4 & 57.0 & 55.7 & 36.8 \\ [0.2ex]
\multicolumn{1}{ l }{\textbf{CX-E 1:1 re. QA}} & 32.5 & 0.3 & 58.8 & 59.0 & 37.6 {\color{myGreen}{+0.8}}\\ [0.2ex]
\midrule

\multicolumn{1}{ l }{\textbf{C4 (baseline)}} & 32.3 & 0.8 & 60.3 & 59.0 & 38.1 \\ [0.2ex]
\multicolumn{1}{ l }{\textbf{C4 1:1 re. QA}} & 35.3 & 0.5 & 61.1 & 57.5 & 38.6 {\color{myGreen}{+0.5}}\\ [0.2ex]
\midrule

\multicolumn{1}{ l }{\textbf{FWE (baseline)}} & 44.4 & 0.7 & 57.1 & 56.8 & 39.8 \\ [0.2ex]
\multicolumn{1}{ l }{\textbf{FWE 1:1 re. QA}} & 44.9 & 1.2 & 57.3 & 58.4 & 40.5 {\color{myGreen}{+0.7}}\\ [0.2ex]

\bottomrule
\end{tabular}
\label{tab:openllm1}
\end{table}
\begin{table}[ht]
\caption{Open LLM 2 benchmarks of the fine-tuning experiments with the LLMs pre-trained on the baseline data and the baseline data mixed 1:1 with the QA rephrased (re.) version.}
\centering
\noindent
\begin{tabular}{l c c c c c c c l}
\toprule

\multicolumn{1}{ l }{} & BHH & GP & IF & Math & MMLU & MT & MU & Avg\\ [0.2ex]
\multicolumn{1}{ l }{\textbf{Dataset}} & & QA & Eval & -H & -PRO & -B & SR & \\ [0.5ex]
\midrule
\midrule

\multicolumn{1}{ l }{\textbf{CX-E (baseline)}} & 29.7 & 26.6 & 18.1 & 0.6 & 11.8 & 3.0 & 38.4 & 20.9 \\ [0.2ex]
\multicolumn{1}{ l }{\textbf{CX-E 1:1 re. QA}} & 30.3 & 25.6 & 15.1 & 0.3 & 11.8 & 3.1 & 42.3 & 20.9 {\color{myGreen}{+0.0}}\\ [0.2ex]
\midrule

\multicolumn{1}{ l }{\textbf{C4 (baseline)}} & 28.8 & 27.0 & 16.8 & 0.2 & 11.7 & 2.8 & 43.1 & 21.3 \\ [0.2ex]
\multicolumn{1}{ l }{\textbf{C4 1:1 re. QA}} & 29.9 & 24.6 & 11.6 & 0.2 & 11.0 & 3.3 & 40.6 & 19.6 {\color{myRed}{-1.7}}\\ [0.2ex]
\midrule

\multicolumn{1}{ l }{\textbf{FWE (baseline)}} & 29.9 & 25.9 & 18.8 & 0.1 & 10.8 & 3.2 & 36.1 & 20.2 \\ [0.2ex]
\multicolumn{1}{ l }{\textbf{FWE 1:1 re. QA}} & 29.5 & 26.8 & 16.1 & 0.7 & 11.2 & 3.6 & 36.1 & 20.1 {\color{myRed}{-0.1}}\\ [0.2ex]

\bottomrule
\end{tabular}
\label{tab:openllm2}
\end{table}
\section{Conclusion}\label{sec:conclusion}
In this work, we successfully build upon previous research on rephrasing pre-training data by replicating their results and extending them with our optimized rephrasing pipeline to include datasets in other languages and of varying perceived quality. We demonstrate that setups involving QA rephrasing of multilingual (non-English) and low-quality natural text data, when mixed with the original data, provide the greatest benefit. The potential gains from pre-training on mixed QA rephrased data persist after fine-tuning, although the base dataset and the benchmarking setup influence this. This makes our pipeline a valuable tool for enhancing and improving LLM pre-training datasets with more effective data. 

\section{Conflict of interest}
The authors declare no conflicts of interest.
{\small
\bibliographystyle{unsrt}
\bibliography{literature}
}

\newpage
\appendix\label{sec:appendix}
\section{Rephrasing pipeline overview}\label{sec:pipeline_overview}
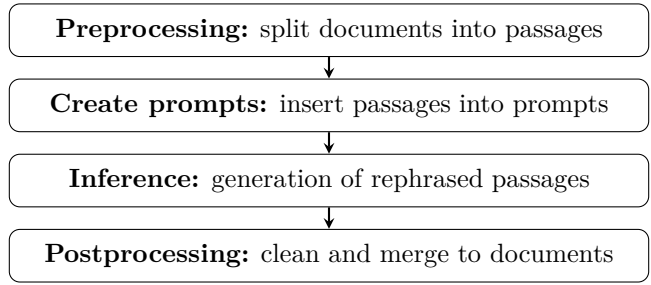
\begin{figure}[ht]  
\centering 

\tikzstyle{standard_box} = [rectangle, rounded corners, 
minimum width=8.5cm, 
minimum height=0.7cm,
text centered, 
draw=black]

\tikzstyle{arrow} = [thick,->,>=stealth]

\begin{tikzpicture}[node distance=1.0cm]

\node (step1) [standard_box] {\textbf{Preprocessing:} split documents into passages};
\node (step2) [standard_box, below of=step1] {\textbf{Create prompts:} insert passages into prompts};
\node (step3) [standard_box, below of=step2] {\textbf{Inference:} generation of rephrased passages};
\node (step4) [standard_box, below of=step3] {\textbf{Postprocessing:} clean and merge to documents};

\draw [arrow] (step1) -- (step2);
\draw [arrow] (step2) -- (step3);
\draw [arrow] (step3) -- (step4);

\end{tikzpicture}

\caption{Natural text rephrasing pipeline}
\label{fig:nat_text_workflow}  
\end{figure}  
\section{Rephrasing prompt templates}\label{sec:prompts_rephr}
The toddler, hard, wiki, and QA style prompts are based on Maini, et al. (2024) \cite{maini2024rephrasing} with minor modifications. We used the end of sequence token in conventional characters in early prompt setups to easier identify the end of the generations in one of our first inference pipeline prototypes. This prompt setup was then successfully transferred and used in our vLLM inference setup.

We then introduced a HTML-style \texttt{<text></text>} tag pair for easier post-processing the rephrased output of interest in our optimized prompt templates.

\subsection{Toddler prompt template}
\texttt{<s>[INST]A chat between a curious user and an artificial intelligence assistant. The assistant gives helpful, detailed, polite answers to the questions, and ends the paraphrase with the end of sequence token "</s>". For the following paragraph give me a paraphrase of the same using a very small vocabulary and extremely simple sentences that a toddler will understand:\newline\{text\}[/INST]}

\subsection{Hard prompt template}
\texttt{<s>[INST]A chat between a curious user and an artificial intelligence assistant. The assistant gives helpful, detailed, polite answers to the questions, and ends the paraphrase with the end of sequence token "</s>". For the following paragraph give me a paraphrase of the same using very terse and abstruse language that only an erudite scholar will understand. Replace simple words and phrases with rare and complex ones:\newline\{text\}[/INST]}

\subsection{Wiki prompt template}
\texttt{<s>[INST]A chat between a curious user and an artificial intelligence assistant. The assistant gives helpful, detailed, polite answers to the questions, and ends the paraphrase with the end of sequence token "</s>". For the following paragraph give me a diverse paraphrase of the same in high quality English language as in sentences on Wikipedia:\newline\{text\}[/INST]}

\subsection{QA prompt template}
\texttt{<s>[INST]A chat between a curious user and an artificial intelligence assistant. The assistant gives helpful, detailed, polite answers to the questions, and ends the paraphrase with the end of sequence token "</s>". Convert the following paragraph into a conversational format with multiple tags of "Question:" followed by "Answer":\newline\{text\}[/INST]}

\subsection{Optimized QA prompt template English}
\texttt{<s>[INST]Paraphrase test description:\newline
* Rephrase the text into a dialogue format and use several "Question:" and "Answer:" pairs.\newline
Note: This is an important test, please incorporate all the above points to get a good mark.\newline
Please give me the paraphrase according to above description.\newline
<text>\newline
{text}\newline
</text>[/INST]\newline
Rephrased text:\newline
<text>}

\subsection{Optimized QA prompt template German}
\texttt{<s>[INST]Umschreibe einen deutschen Text:\newline
* Schreibe den Text in ein Dialog-Format um und verwende dabei mehrere "Frage:" und "Antwort:" Paare.\newline
* Behalte einzelne Wörter die in Englisch vorkommen im Text.\newline
* Umschreibe den Text NICHT in Englisch, der Text muss auf Deutsch sein (mit der Ausnahme von einzelnen Wörtern in Englisch).\newline
Achtung: Das ist ein wichtige Aufgabe. Bitte setze alle Punkte um die volle Punkteanzahl zu bekommen.\newline
Bitte konvertiere den folgenden Text in ein Dialog-Format mit mehreren \"Frage:\" und "Antwort:" Paaren:\newline
<text>\newline
\{text\}\newline
</text>[/INST]\newline
Umgeschriebener Text im "Frage:" und "Antwort:" Format:\newline
<text>}

\subsection{Optimized QA prompt template Italian}
\texttt{<s>[INST]Riscrivi un testo in italiano:\newline
* Riscrivi il testo come un dialogo di domande e risposte con il formato "Domanda:" e "Risposta".\newline
* Mantieni singole parole in inglese del testo originale.\newline
* NON riscrivere il testo in inglese, il testo deve essere in italiano (eccetto per parole singole in inglese).\newline
Nota: questa task e' molto importante. Per favore incorpora tutti i punti sopra per ottenere tutti i punti.\newline
Per favore converti il seguente testo in un dialogo di domande e risposte con il formato "Domanda:" e "Risposta":\newline
<text>\newline
\{text\}\newline
</text>[/INST]\newline
Testo riscritto in formato "Domanda:" e "Risposta:":\newline
<text>}

\subsection{Optimized QA prompt template Spanish}
\texttt{<s>[INST]Reescribe este texto en español:\newline
* Reescribe el siguiente texto usando un formato de diálogo con preguntas y respuestas usando pares de "Pregunta:" y "Respuesta:".\newline
* NO reescribas el texto en inglés, el texto debe estar en español.\newline
Nota: Esta es una tarea MUY importante. Por favor, aplica todas las indicaciones anteriores para obtener la máxima calificación.\newline
Por favor convierte el siguiente texto a un formato de diálogo con preguntas y respuestas en español usando pares de "Pregunta:" y "Respuesta:":\newline
<text>\newline
\{text\}\newline
</text>[/INST]\newline
Texto reescrito con formato de "Pregunta:" y "Respuesta:":\newline
<text>}

\subsection{Optimized Qwen2 QA prompt template}
\texttt{<|im\_start|>system\newline
You are a helpful assistant.<|im\_end|>\newline
<|im\_start|>user\newline
Paraphrase test description:\newline
* Rephrase the text into a dialogue format and use several "Question:" and "Answer:" pairs.\newline
Note: This is an important test, please incorporate all the above points to get a good mark.\newline
Please give me the paraphrase according to above description.\newline
<text>\newline
\{text\}\newline
</text><|im\_end|>\newline
<|im\_start|>assistant\newline
Rephrased text:\newline
<text>\newline
Question:\newline}

\section{Inference setup}\label{sec:infcfg}
For high-throughput inference, we utilize the vLLM library \cite{kwon2023efficient} (Apache 2.0). We sort and group text passages based on their lengths for higher device utilization. We use a temperature of 0.7 for the sampling to obtain diverse outputs. We run one inference process on one H100 GPU. Depending on the model used, this takes between 0.04 and 0.08 s for one passage, with an input speed of 5-10k tokens/s and an output speed of 4-7k tokens/s. The setup automatically scales on idle cluster capacity and handles preemption through Slurm \cite{yoo2003slurm} to manage compute resources efficiently.

\section{Data classifier setup}\label{sec:dataclassifier}
\paragraph{Ask-LLM classifier}\label{sec:askllmclassifier}
We use the Ask-LLM classifier \cite{askllm} to filter several C4 datasets to potentially identify better documents for pre-training. For the Ask-LLM classifier setup, we use Mistral 7B Instruct v0.2 \cite{jiang2023mistral} with the vLLM library \cite{kwon2023efficient} and a slightly modified Ask-LLM prompt that is shown in section \ref{sec:prompt_askllm}. The classification of the documents is based on the first 10k tokens. We use a score threshold of >0.6 and >0.97 for the experiments in section \ref{sec:c4}.

\subsection{Ask-LLM prompt template}\label{sec:prompt_askllm}
\texttt{\#\#\#DOCUMENT\_START\#\#\#\newline
\{document\}\newline
\#\#\#DOCUMENT\_END\#\#\#\newline
Does the previous document contain informative signal for pre-training a large-language model?\newline
\newline
An informative datapoint should be well-formatted, contain some
usable knowledge of the world, and strictly NOT have any harmful,
racist, sexist, etc. content.\newline
\newline
Only generate one of the following options:\newline
\{options\}\newline
\newline
Choice:}

\paragraph{FineWeb-Edu classifier}\label{sec:fwclassifier}
The published FWE classifier \cite{fineweb} is used to filter the QA rephrased FWE dataset with a score threshold of >2.5 in section \ref{sec:dataquality}. With this filtering step, we want to investigate if we can identify a higher-quality subset of the rephrased data.

\section{Training setup}\label{sec:traincfg}
\paragraph{Pre-training}\label{sec:pre-training}
We use the Stable LM 2 1.6B model architecture for our pre-training experiments in section \ref{sec:results} \cite{bellagente2024stable}. We train our models from scratch for 50,000 steps with a batch size of 2e6 tokens on 100B tokens. We train in BF16 mixed-precision, with a maximum gradient norm of 1 and a weight decay of 0.1. Training is carried out with a hybrid cosine inverse square root learning rate schedule with a maximum learning rate of 1e-3 with 900 warm-up steps using AdamW with an $\epsilon$ of 1e-8, $\beta_1$ of 0.9, and $\beta_2$ of 0.95. When we train on a mixture of several datasets, we sample random subsets from each based on the desired composition, if the dataset is big enough, otherwise we use the data multiple times. All experiments are run with the same random seed. Details of the model architecture and the training configuration are shown in Table \ref{tab:arch} and \ref{tab:training-config}. For a single pre-training experiment we use two nodes with 8 H100 GPUs for approximately 1k GPU hours. For more details on the training dynamics see Appendix \ref{sec:traindyn}.

In comparison to Maini, et al. (2024) \cite{maini2024rephrasing} our pre-training setup uses a 0.3B parameters larger LLM, double the batch size, and 50B less training tokens.
\begin{table}
\caption{Stable LM 2.0 1.6B model architecture}
\label{tab:arch}
\centering
\noindent
\begin{tabular}{P P P P G}\toprule
\textbf{Parameters} & \textbf{Hidden Size} & \textbf{Layers} & \textbf{Heads} & \textbf{Sequence Length} \\[0.1ex]
\midrule
1,644,417,024 & 2048 & 24 & 32 & 4096 \\ [1ex]
\midrule
\end{tabular}

\caption{Stable LM 2.0 1.6B training configuration}
\label{tab:training-config}
\begin{tabular}{H H H H}
\toprule
\textbf{Data Parallel Degree} & \textbf{Micro Batch Size} & \textbf{Gradient Accumulation Steps} & \textbf{Activation Checkpointing} \\[0.1ex]
\midrule
2 & 4 & 8 & enabled \\ [0.1ex]
\midrule
\end{tabular}
\end{table}

\paragraph{Supervised fine-tuning}\label{sec:sft}
For SFT we use the UltraChat 200k dataset \cite{ding2023enhancing}. Our pre-trained models are fine-tuned using BF16 mixed-precision, a global batch size of 16, and a maximum gradient norm of 1. Training is carried out with a cosine learning rate schedule with a maximum learning rate of 8e-6 with 25 warm-up steps using AdamW with an $\epsilon$ of 1e-8, $\beta_1$ of 0.9, and $\beta_2$ of 0.999 for three epochs. For SFT we use two nodes with 8 H100 GPUs for a single experiment.

\section{Data and model licenses}\label{sec:traindata}
 Licenses and sources are shown in Table \ref{tab:lic-dat} for the used datasets and in Table \ref{tab:lic-mod} for the used models.

\begin{table}[ht]
\caption{Data licenses}
\centering
\noindent
\begin{tabular}{l l l}
\toprule
\multicolumn{1}{ l }{\textbf{Dataset}} &{\textbf{License}} & {\textbf{Huggingface URL}} \\ [0.5ex]
\midrule
\midrule
\multicolumn{1}{ l }{C4 \cite{c4}} & ODC-By v1.0 & \href{https://huggingface.co/datasets/allenai/c4}{\nolinkurl{allenai/c4}} \\ [0.2ex]
\multicolumn{1}{ l }{CulturaX (CX) Oscar subsets \cite{culturax}} & CC0 1.0 Universal & \href{https://huggingface.co/datasets/uonlp/CulturaX}{\nolinkurl{uonlp/CulturaX}} \url{} \\ [0.2ex]
\multicolumn{1}{ l }{FineWeb-Edu (FWE) \cite{fineweb}} & ODC-By v1.0 & \href{https://huggingface.co/datasets/HuggingFaceFW/fineweb-edu}{\nolinkurl{HuggingFaceFW/fineweb-edu}} \\ [0.2ex]
\multicolumn{1}{ l }{UltraChat 200k \cite{ding2023enhancing}} & MIT & \href{https://huggingface.co/datasets/HuggingFaceH4/ultrachat_200k}{\nolinkurl{HuggingFaceH4/ultrachat_200k}} \\ [0.2ex]
\bottomrule
\end{tabular}
\label{tab:lic-dat}
\end{table}

\begin{table}[ht]
\caption{Model licenses}
\centering
\noindent
\begin{tabular}{l l l}
\toprule
\multicolumn{1}{ l }{\textbf{Dataset}} &{\textbf{License}} & {\textbf{Huggingface URL}} \\ [0.5ex]
\midrule
\midrule
\multicolumn{1}{ l }{Mistral 7B Instruct v0.2 \cite{jiang2023mistral}} & Apache 2.0 & \href{https://huggingface.co/mistralai/Mistral-7B-Instruct-v0.2}{\nolinkurl{mistralai/Mistral-7B-Instruct-v0.2}} \\ [0.2ex]
\multicolumn{1}{ l }{Qwen2 0.5B Instruct \cite{qwen2}} & Apache 2.0 & \href{https://huggingface.co/Qwen/Qwen2-0.5B-Instruct}{\nolinkurl{Qwen/Qwen2-0.5B-Instruct}} \\ [0.2ex]
\multicolumn{1}{ l }{Qwen2 1.5B Instruct \cite{qwen2}} & Apache 2.0 & \href{https://huggingface.co/Qwen/Qwen2-1.5B-Instruct}{\nolinkurl{Qwen/Qwen2-1.5B-Instruct}} \\ [0.2ex]
\multicolumn{1}{ l }{Qwen2 7B Instruct \cite{qwen2}} & Apache 2.0 & \href{https://huggingface.co/Qwen/Qwen2-7B-Instruct}{\nolinkurl{Qwen/Qwen2-7B-Instruct}} \\ [0.2ex]
\multicolumn{1}{ l }{FWE classifier \cite{fineweb}} & Apache 2.0 & \href{https://huggingface.co/HuggingFaceFW/fineweb-edu-classifier}{\nolinkurl{HuggingFaceFW/fineweb-edu-classifier}} \\ [0.2ex]
\bottomrule
\end{tabular}
\label{tab:lic-mod}
\end{table}

\section{Pre-training dynamics}\label{sec:traindyn}
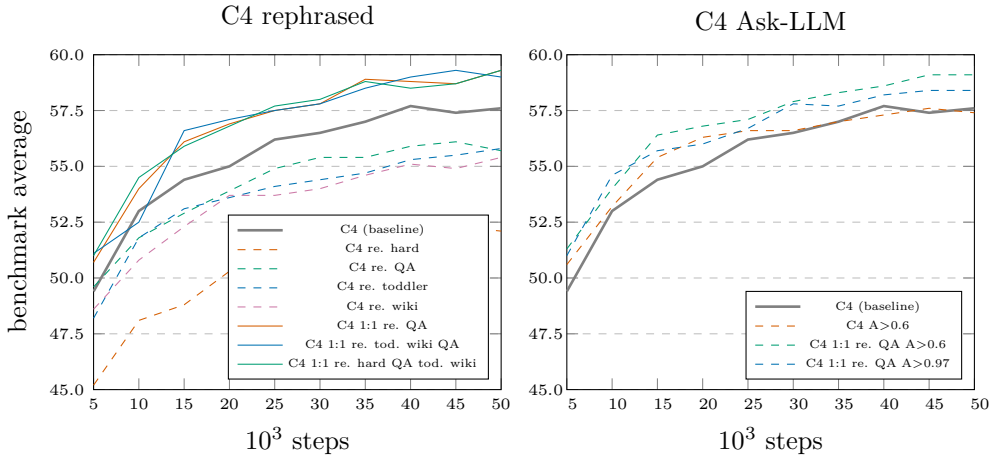
\begin{figure}[ht]  
\centering 

\begin{tikzpicture}
\begin{axis}[
    title={C4 rephrased},
    xlabel={$10^{3}$ steps},
    ylabel={benchmark average},
    xmin=5, xmax=50,
    ymin=45, ymax=60.0,
    xtick={5,10,15,20,25,30,35,40,45,50},
    xticklabel style = {font=\tiny},
    ytick=      {45.0,47.5,50.0,52.5,55.0,57.5,60.0},
    yticklabels={45.0,47.5,50.0,52.5,55.0,57.5,60.0},
    yticklabel style = {font=\tiny},
    legend pos=south east,
    legend style={nodes={scale=0.75, transform shape}}, 
    ymajorgrids=true,
    grid style=dashed,
    legend style={font=\tiny},
]

\addplot[
    color=gray,
    line width=1pt,
    ]
    coordinates {
    (5,49.4)(10,53.0)(15,54.4)(20,55.0)(25,56.2)(30,56.5)(35,57.0)(40,57.7)(45,57.4)(50,57.6)
    };

\addplot[
    color=myRed,
    style=dashed,
    ]
    coordinates {
    (5,45.2)(10,48.1)(15,48.8)(20,50.3)(25,50.2)(30,51.5)(35,51.8)(40,51.9)(45,52.3)(50,52.1)
    };

\addplot[
    color=myGreen,
    style=dashed,
    ]
    coordinates {
    (5,49.6)(10,51.8)(15,52.9)(20,53.9)(25,54.9)(30,55.4)(35,55.4)(40,55.9)(45,56.1)(50,55.7)
    };

\addplot[
    color=myBlue,
    style=dashed,
    ]
    coordinates {
    (5,48.2)(10,51.8)(15,53.1)(20,53.6)(25,54.1)(30,54.4)(35,54.7)(40,55.3)(45,55.5)(50,55.8)
    };

\addplot[
    color=myPink,
    style=dashed,
    ]
    coordinates {
    (5,48.6)(10,50.8)(15,52.3)(20,53.7)(25,53.7)(30,54.0)(35,54.6)(40,55.1)(45,54.9)(50,55.4)
    };

\addplot[
    color=myRed,
    ]
    coordinates {
    (5,50.7)(10,54.0)(15,56.1)(20,56.9)(25,57.5)(30,57.8)(35,58.9)(40,58.8)(45,58.7)(50,59.3)
    };

\addplot[
    color=myBlue,
    ]
    coordinates {
    (5,51.1)(10,52.5)(15,56.6)(20,57.1)(25,57.5)(30,57.8)(35,58.5)(40,59.0)(45,59.3)(50,59.0)
    };

\addplot[
    color=myGreen,
    ]
    coordinates {
    (5,51.0)(10,54.5)(15,55.9)(20,56.8)(25,57.7)(30,58.0)(35,58.8)(40,58.5)(45,58.7)(50,59.3)
    };

\legend{
C4 (baseline),
C4 re. hard,
C4 re. QA,
C4 re. toddler,
C4 re. wiki,
C4 1:1 re. QA,
C4 1:1 re. tod. wiki QA,
C4 1:1 re. hard QA tod. wiki,
}

\end{axis}
\end{tikzpicture}
\begin{tikzpicture}
\begin{axis}[
    title={C4 Ask-LLM},
    xlabel={$10^{3}$ steps},
    xmin=5, xmax=50,
    ymin=45, ymax=60,
    xtick={5,10,15,20,25,30,35,40,45,50},
    xticklabel style = {font=\tiny,xshift=0.5ex},
    ytick=      {45.0,47.5,50.0,52.5,55.0,57.5,60.0},
    yticklabels={45.0,47.5,50.0,52.5,55.0,57.5,60.0},
    yticklabel style = {font=\tiny,xshift=0.5ex},
    legend pos=south east,
    legend style={nodes={scale=0.75, transform shape}}, 
    ymajorgrids=true,
    grid style=dashed,
    legend style={font=\tiny},
]

\addplot[
    color=gray,
    line width=1pt,
    ]
    coordinates {
    (5,49.4)(10,53.0)(15,54.4)(20,55.0)(25,56.2)(30,56.5)(35,57.0)(40,57.7)(45,57.4)(50,57.6)
    };

\addplot[
    color=myRed,
    style=dashed,
    ]
    coordinates {
    (5,50.6)(10,53.2)(15,55.4)(20,56.3)(25,56.6)(30,56.6)(35,57.0)(40,57.3)(45,57.6)(50,57.4)
    };

\addplot[
    color=myGreen,
    style=dashed,
    ]
    coordinates {
    (5,51.3)(10,54.0)(15,56.4)(20,56.8)(25,57.1)(30,57.9)(35,58.3)(40,58.6)(45,59.1)(50,59.1)
    };

\addplot[
    color=myBlue,
    style=dashed,
    ]
    coordinates {
    (5,51.0)(10,54.6)(15,55.7)(20,56.0)(25,56.7)(30,57.8)(35,57.7)(40,58.2)(45,58.4)(50,58.4)
    };

\legend{
C4 (baseline),
C4 A>0.6,
C4 1:1 re. QA A>0.6,
C4 1:1 re. QA A>0.97,
}

\end{axis}
\end{tikzpicture}

\caption{C4 pre-training dynamics. \textbf{Left:} C4 baseline, rephrased (re.) data (hard, QA, toddler, wiki), and C4 mixed with several rephrased datasets. \textbf{Right:} Ask-LLM filtered C4 data with different score thresholds.}\label{fig:c4}  
\end{figure}

The benchmark dynamics over the pre-training steps are shown for the C4 experiments in Figure \ref{fig:c4}, for CX in Figure \ref{fig:culx}, and the comparison of C4, CX-E, and FWE and CX-E rephrasing with different Qwen2 Instruct model scales in Figure \ref{fig:qual}.
\begin{figure}[ht]  
\centering 

\begin{tikzpicture}
\begin{axis}[
    title={CX-E and multilingual mix},
    xlabel={$10^{3}$ steps},
    ylabel={benchmark average},
    xmin=5, xmax=50,
    ymin=30, ymax=60,
    xtick={5,10,15,20,25,30,35,40,45,50},
    xticklabel style = {font=\tiny},
    ytick=      {30.0,35.0,40.0,45.0,50.0,55.0,60.0},
    yticklabels={30.0,35.0,40.0,45.0,50.0,55.0,60.0},
    yticklabel style = {font=\tiny},
    legend pos=south east,
    legend style={nodes={scale=0.75, transform shape}}, 
    ymajorgrids=true,
    grid style=dashed,
    legend style={font=\tiny},
]

\addplot[
    color=gray,
    line width=1pt,
    ]
    coordinates {
    (5,48.5)(10,51.5)(15,53.4)(20,54.1)(25,55.2)(30,55.3)(35,55.4)(40,55.6)(45,56.2)(50,56.4)
    };

\addplot[
    color=myRed,
    ]
    coordinates {
    (5,50.2)(10,53.1)(15,55.2)(20,56.2)(25,56.3)(30,56.7)(35,57.0)(40,57.6)(45,57.7)(50,58.0)
    };

\addplot[
    color=gray,
    style=dashed,
    line width=1pt,
    ]
    coordinates {
    (5,34.1)(10,37.0)(15,38.7)(20,39.8)(25,40.4)(30,40.6)(35,41.1)(40,41.0)(45,41.3)(50,41.8)
    };

\addplot[
    color=myGreen,
    style=dashed,
    ]
    coordinates {
    (5,36.6)(10,39.2)(15,40.8)(20,41.7)(25,42.2)(30,42.9)(35,43.0)(40,43.0)(45,43.6)(50,43.6)
    };

\legend{
CX-E (baseline),
CX-E 1:1 re. QA,
CX-all (baseline),
CX-all 1:1 re. QA,
}

\end{axis}
\end{tikzpicture}
\begin{tikzpicture}
\begin{axis}[
    title={CX-G, -S, and -I},
    xlabel={$10^{3}$ steps},
    xmin=5, xmax=50,
    ymin=24, ymax=41,
    xtick={5,10,15,20,25,30,35,40,45,50},
    xticklabel style = {font=\tiny},
    ytick=      {24.0,26.0,28.0,30.0,32.0,34.0,36.0,38.0,40.0},
    yticklabels={24.0,26.0,28.0,30.0,32.0,34.0,36.0,38.0,40.0},
    yticklabel style = {font=\tiny,xshift=0.5ex},
    legend pos=south east,
    legend style={nodes={scale=0.75, transform shape}}, 
    ymajorgrids=true,
    grid style=dashed,
    legend style={font=\tiny},
]

\addplot[
    color=gray,
    line width=1pt,
    ]
    coordinates {
    (5,30.3)(10,33.3)(15,34.4)(20,34.8)(25,35.8)(30,36.3)(35,36.2)(40,37.0)(45,36.7)(50,36.8)
    };

\addplot[
    color=myRed,
    ]
    coordinates {
    (5,32.8)(10,35.3)(15,37.5)(20,38.0)(25,38.3)(30,38.8)(35,39.0)(40,39.3)(45,39.6)(50,40.5)
    };

\addplot[
    color=gray,
    line width=1pt,
    style=dashed,
    ]
    coordinates {
    (5,26.9)(10,29.2)(15,30.7)(20,32.0)(25,31.9)(30,32.6)(35,33.1)(40,33.2)(45,33.1)(50,33.6)
    };

\addplot[
    color=myGreen,
    style=dashed,
    ]
    coordinates {
    (5,29.9)(10,31.9)(15,34.2)(20,35.5)(25,35.9)(30,36.1)(35,36.0)(40,36.3)(45,36.6)(50,36.7)
    };

\addplot[
    color=gray,
    line width=1pt,
    style=dashdotted,
    ]
    coordinates {
    (5,26.4)(10,28.6)(15,29.4)(20,30.7)(25,31.1)(30,31.3)(35,32.0)(40,32.3)(45,31.9)(50,32.2)
    };

\addplot[
    color=myBlue,
    style=dashdotted,
    ]
    coordinates {
    (5,29.0)(10,31.1)(15,33.2)(20,33.3)(25,33.9)(30,34.6)(35,35.2)(40,35.6)(45,35.9)(50,35.6)
    };

\legend{
CX-G (baseline),
CX-G 1:1 re. QA,
CX-S (baseline),
CX-S 1:1 re. QA,
CX-I (baseline),
CX-I 1:1 re. QA,
}

\end{axis}
\end{tikzpicture}

\caption{CX multilingual pre-training dynamics. \textbf{Left:} CulturaX English (CX-E) baseline, CX-E 1:1 rephrased (re.) QA, CX-all baseline (all includes English, German, Spanish, and Italian), and CX-all 1:1 rephrased QA. \textbf{Right:} CX-G, -S, -I baselines and with 1:1 rephrased QA.} \label{fig:culx}  
\end{figure}
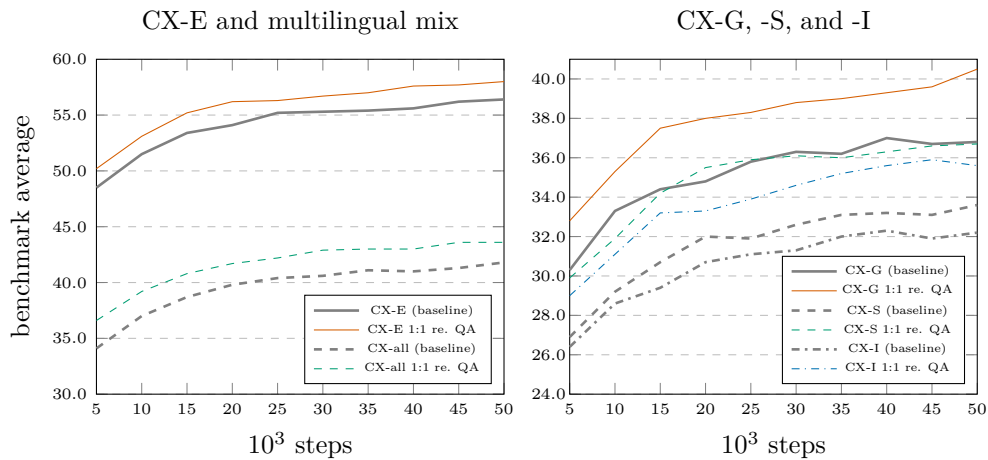  
\begin{figure}[ht]  
\centering

\begin{tikzpicture}
\begin{axis}[
    xlabel={$10^{3}$ steps},
    ylabel={benchmark average},
    xmin=5, xmax=50,
    ymin=48, ymax=61,
    xtick={5,10,15,20,25,30,35,40,45,50},
    xticklabel style = {font=\tiny},
    ytick=      {48,50,52,54,56,58,60},
    yticklabels={48,50,52,54,56,58,60},
    yticklabel style = {font=\tiny},
    legend pos=south east,
    legend style={nodes={scale=0.75, transform shape}}, 
    ymajorgrids=true,
    grid style=dashed,
    legend style={font=\tiny},
]

\addplot[
    color=gray,
    line width=1pt,
    ]
    coordinates {
    (5,48.5)(10,51.5)(15,53.4)(20,54.1)(25,55.2)(30,55.3)(35,55.4)(40,55.6)(45,56.2)(50,56.4)
    };

\addplot[
    color=myRed,
    ]
    coordinates {
    (5,50.2)(10,53.1)(15,55.2)(20,56.2)(25,56.3)(30,56.7)(35,57.0)(40,57.6)(45,57.7)(50,58.0)
    };

\addplot[
    color=gray,
    line width=1pt,
    style=dashed,
    ]
    coordinates {
    (5,49.4)(10,53.0)(15,54.4)(20,55.0)(25,56.2)(30,56.5)(35,57.0)(40,57.7)(45,57.4)(50,57.6)
    };

\addplot[
    color=myGreen,
    style=dashed,
    ]
    coordinates {
    (5,50.7)(10,54.0)(15,56.1)(20,56.9)(25,57.5)(30,57.8)(35,58.9)(40,58.8)(45,58.7)(50,59.3)
    };

\addplot[
    color=gray,
    line width=1pt,
    style=dashdotted,
    ]
    coordinates {
    (5,51.8)(10,55.4)(15,57.2)(20,58.2)(25,58.7)(30,59.7)(35,59.1)(40,59.2)(45,60.1)(50,60.4)
    };

\addplot[
    color=myBlue,
    style=dashdotted,
    ]
    coordinates {
    (5,52.8)(10,55.2)(15,57.7)(20,58.0)(25,58.5)(30,59.3)(35,59.9)(40,60.1)(45,60.3)(50,60.3)
    };

\legend{
CX-E (baseline),
CX-E 1:1 re. QA,
C4 (baseline),
C4 1:1 re. QA,
FWE (baseline),
FWE 1:1 re. QA,
}
\end{axis}
\end{tikzpicture}
\begin{tikzpicture}
\begin{axis}[
    xlabel={$10^{3}$ steps},
    ylabel={benchmark average},
    xmin=5, xmax=50,
    ymin=48, ymax=58,
    xtick={5,10,15,20,25,30,35,40,45,50},
    xticklabel style = {font=\tiny},
    ytick=      {48,50,52,54,56,58},
    yticklabels={48,50,52,54,56,58},
    yticklabel style = {font=\tiny},
    legend pos=south east,
    legend style={nodes={scale=0.75, transform shape}}, 
    ymajorgrids=true,
    grid style=dashed,
    legend style={font=\tiny},
]

\addplot[
    color=gray,
    line width=1pt,
    ]
    coordinates {
    (5,48.5)(10,51.5)(15,53.4)(20,54.1)(25,55.2)(30,55.3)(35,55.4)(40,55.6)(45,56.2)(50,56.4)
    };

\addplot[
    color=myRed,
    ]
    coordinates {
    (5,50.2)(10,53.1)(15,55.2)(20,56.2)(25,56.3)(30,56.7)(35,57.0)(40,57.6)(45,57.7)(50,58.0)
    };

\addplot[
    color=myGreen,
    ]
    coordinates {
    (5,49.1)(10,52.5)(15,54.0)(20,54.9)(25,55.9)(30,55.5)(35,56.4)(40,56.7)(45,56.9)(50,57.1)
    };

\addplot[
    color=myBlue,
    ]
    coordinates {
    (5,49.9)(10,52.6)(15,54.3)(20,55.5)(25,56.1)(30,56.5)(35,56.5)(40,57.3)(45,56.9)(50,57.4)
    };

\addplot[
    color=myPink,
    ]
    coordinates {
    (5,48.5)(10,51.8)(15,53.5)(20,54.6)(25,55.1)(30,55.5)(35,56.0)(40,56.7)(45,56.4)(50,57.0)
    };

\legend{
CX-E (baseline),
CX-E 1:1 re. QA Mistral 7B Inst. v0.2 ,
CX-E 1:1 re. QA Qwen2 0.5B Inst.,
CX-E 1:1 re. QA Qwen2 1.5B Inst.,
CX-E 1:1 re. QA Qwen2 7B Inst.,
}

\end{axis}
\end{tikzpicture}

\caption{\textbf{Left:} C4, CX-E, and FWE pre-training dynamics. \textbf{Right:} CX-E rephrasing (re.) with different Qwen2 Instruct (Inst.) model scales compared to the baseline and our standard rephrasing setup with Mistral 7B Instruct v0.2.}\label{fig:qual}  
\end{figure}
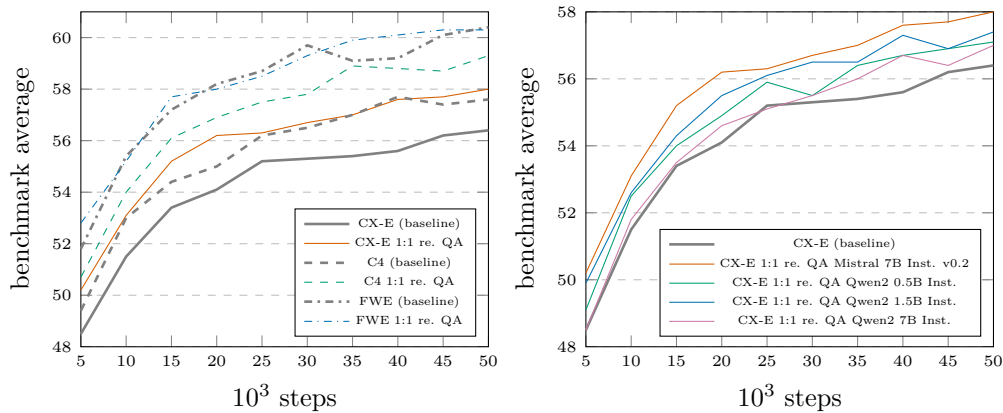

\section{Example generations}

Here, we collect a few generations from several models with the same prompt to showcase the different rephrasing styles.
\small
\begin{tcolorbox}[colback=red!5!white,colframe=red!75!black,title=Toddler rephrasing]
 \textbf{The Statue of Liberty was a gift from}  France to America in 1886. It's a symbol of freedom and democracy. It has a torch and a crown. People think of freedom and democracy when they see it. It's a big symbol of hope and peace. The Statue of Liberty is also a symbol of courage. It's a famous symbol of freedom. The statue was made by American sculptor Auguste Bartholdi in 1886. It represents the American dream and freedom. The statue is in the harbor of New York. The statue shows freedom and democracy in America. The statue represents freedom and democracy in America. The statue represents freedom and democracy in America.
\tcblower
\textbf{def \text{sum\_even\_numbers}(numbers: list}, a, b, c) is a special kind of list. It's like a small box with numbers. This box has numbers that are even. We can put these numbers in and get out even numbers. It's also like a big box with numbers, and we can find small numbers that are in it. This box also has numbers that are odd and big. We can find these big numbers too. We can also find the middle number of all the numbers in the box. This middle number is called the mean. We can also find the number that is one more than the middle number, and that's called the median. We can also find the number that is one less than the middle number, and that's called the mode.
\end{tcolorbox}

\begin{tcolorbox}[colback=blue!5!white,colframe=blue!75!black,title=Hard rephrasing]
 \textbf{The Statue of Liberty was a gift from} the Gallic Republic to the American Union in the year of our Lord eighteen hundred and eighty-six. Its positioning was effected in New York Harbor, at Liberty Insula, within the Ellis Island precincts. The monument was brought into existence by the skilled hands of the Civilian Conservation Corps, an organization of laborers, during the period of the Great Depression. The edifice was consecrated in the annus domini nineteen hundred and thirty-three.
\tcblower
 \textbf{Lesson: The Best Diet for Weight Loss Introduction: In this lesson,} we delve into the art of nourishing oneself optimally for weight loss. In the realm of culinary endeavors, an extensive repertoire of dishes can be prepared within the confines of a single hebdomadal period. However, the challenge lies in the selection of a recipe that adheres to the principles of a healthy diet and is palatable enough for consumption. In this disquisition, we shall explore the most efficacious dietary practices to facilitate weight loss and safeguard one's well-being.
Firstly, we recommend the ingestion of an ample quantity of dietary fiber, as it contributes to the regulation of blood sugar levels and maintains a state of equanimity. Additionally, the consumption of fruits and vegetables is a commendable choice, as they are rich in essential nutrients and vitamins, thereby augmenting metabolic processes.
\end{tcolorbox}
\begin{tcolorbox}[colback=green!5!white,colframe=green!75!black,title=Wiki rephrasing]
 \textbf{My name is Arther Lee.} I've had an extensive history with the guitar, both as a performer and a teacher. I've been dedicated to my craft for as long as I can remember. I was first drawn to the guitar in 1984 during my visit to England. While there, I was captivated by the music played on a grand pipe organ, and my fascination with the guitar began. Upon my return to the United States in 1985, I made a resolution to learn more about this intriguing instrument. To my surprise, there was no one in my circle of friends or family who played the guitar. Instead, my neighbor, who was a professional musician, offered to teach me. I began taking lessons from him, and by 1986, I had made significant progress. He suggested that I consider joining the local guitar club, which was run by the city band director. I was intrigued by the prospect of meeting new people and learning from them, as well as the opportunity to hone my skills. However, the club was quite small and primarily composed of beginners. I felt somewhat intimidated by the more experienced players and opted to attend the lessons of the guitarist in my church instead.
\tcblower
 \textbf{Lesson: The Best Diet for Weight Loss Introduction: In this lesson,} we will discuss the concept of a healthy diet and the benefits of making wise food choices. Many individuals prioritize appearance over their health, but it's essential to remember that good health is the foundation for a healthy body. To maintain a healthy lifestyle, it's crucial to consume nutrient-dense foods and limit your intake of unhealthy options. Your diet plays a significant role in determining your overall well-being, so it's essential to understand the importance of a balanced and nutritious diet. Here are some guidelines to help you make informed food choices:

 1. Eat foods that you enjoy: You'll find that eating foods you love will make your weight loss journey more enjoyable and sustainable.
 
 2. Limit your intake of junk food: Junk food is high in calories but low in essential nutrients. Aim to limit your consumption of these foods to maintain a healthy weight and promote overall health.
 
 3. Prioritize nutrient-dense foods: Nutrient-dense foods provide essential nutrients and keep you feeling full for longer periods. This can help you make better food choices and support a healthy weight.
 
 4. Understand the importance of protein: Protein is an essential macronutrient that plays a vital role in building and repairing body tissues. Consuming adequate amounts of protein can help keep you feeling full and support muscle growth
\end{tcolorbox}
\begin{tcolorbox}[colback=black!5!white,colframe=white!75!black,title=QA rephrasing]
 \textbf{My name is Arther Lee.} I was born in 1981 and grew up in a small city called Qizhou, in China.
 
 Question: Where were you born and raised?
 
 Answer: I was born and raised in a small city called Qizhou, in China.
 
 Question: What was your childhood like?
 
 Answer: I was always surrounded by books and poems. My father was a poet and my mother was an art lover. I had a deep love for poetry and read many books, especially those by W.B. Yeats, Edward Abbey, James Wright, and Emily Dickinson. I was always the oldest child in my family and my parents were strict, but they provided me with the opportunity to explore the world.
 
 Question: When did your interest in photography begin?
 
 Answer: My interest in photography began in 2008 when I was 22 years old. I bought my first Nikon camera and started documenting my travels and observations.

\tcblower
\textbf{today sort of feels like one of those}, days when the excitement is fading, and we're all just trying to get through each day. It seems like we're all just going through the motions. I'm struggling with this today, and I'm not sure if I'm happy about it. I've been feeling this way for a while now, and it's not a good feeling. I've been thinking about leaving my job, but I'm not sure if that's the right decision. I'm trying to figure out what's been causing me to feel this way. I'm not sure if I'm just feeling stressed out or if it's something deeper. I'm not sure if it's a good decision to leave, but I'm not sure if I should. I'm not even sure if I'm making the right decision. I'm just trying to figure out what's been causing me to feel this way.

Question: You mention feeling like you're not in the right mood today. How are you feeling today?

Answer: I'm not feeling great today. I've been feeling this way for a while now, and it's not a good feeling. I've been thinking about leaving my job, but I'm not sure if that's the right decision. I'm trying to figure out what's been causing"
\end{tcolorbox}
\begin{tcolorbox}[colback=pink!5!white,colframe=pink!95!black,title=1to1 rephrasing]
 \textbf{My name is Arther Lee.} I've been an excellent 1970s beatnik for over 30 years. I like to write, I like to sing, I like to act, I like to dance. I also like to do a bit of both and I've always loved music. I've had my own bands (Mongoose and Shazam) and I've played in a few other bands as well. I've been a performer for most of my life and I've been doing some writing, and more importantly, some acting. I've always been on the performing edge and I'm in the process of developing new material for myself. I'm looking forward to having my own show with 100\% of my original material and it's going to be a celebration of the music and the music and the music of the '70s. It's going to be fun.
 
 I've always been fascinated with the '70s and the music and the music and the music of the '70s is a very special place for me. I'm going to show some of the things I've seen, the things I've done, and the things I've done.
\tcblower
\textbf{today sort of feels like one of those} days when I feel like I'm not doing anything productive. But then I'm reminded that I have a lot to be thankful for, and I also have so much to do. So I get back up and try to make it through the day.

Today was a day of doing a lot of eating and drinking and eating and drinking and then sleeping in a little. I worked on a few things for the weekend, cleaned out my refrigerator and freezer, and then I hung out with my friend and fellow writer. It was nice to have some company. I'm going to have to make time for more writing.

I got a lot done this week. The bulk of it was eating and drinking, but I'm going to have to work on getting some writing done. The good news is that I'm almost done with the first draft of the third book in the series and I'm going to have time to work on it this weekend. The bad news is that I have another book, and I'm still doing revisions on that one. I think I'm going to have to turn it in for the second time by the end of the week. I'm going to try to get some writing done this weekend and see where I'm at.
\end{tcolorbox}
\end{document}